# Autonomous Urban Localization and Navigation with Limited Information*

Jordan B. Chipka and Mark Campbell

*Abstract*— Urban environments offer a challenging scenario for autonomous driving. Globally localizing information, such as a GPS signal, can be unreliable due to signal shadowing and multipath errors. Detailed *a priori* maps of the environment with sufficient information for autonomous navigation typically require driving the area multiple times to collect large amounts of data, substantial post-processing on that data to obtain the map, and then maintaining updates on the map as the environment changes. This paper addresses the issue of autonomous driving in an urban environment by investigating algorithms and an architecture to enable fully functional autonomous driving with limited information. An algorithm to autonomously navigate urban roadways with little to no reliance on an *a priori* map or GPS is developed. Localization is performed with an extended Kalman filter with odometry, compass, and sparse landmark measurement updates. Navigation is accomplished by a compass-based navigation control law. Key results from Monte Carlo studies show success rates of urban navigation under different environmental conditions. Experiments validate the simulated results and demonstrate that, for given test conditions, an expected range can be found for a given success rate.

I. INTRODUCTION

Autonomous driving, as with many other robotic applications, requires accurate localization to perform robustly. However, dense urban environments provide a key challenge due to the lack of reliable information sources for localization and planning. For example, multipath errors and signal shadowing in dense urban environments make positioning systems based on GPS an unreliable information source for autonomous agents [1]. Even highly accurate, state-of-the-art positioning systems struggle to provide the level of localization needed for autonomous driving due to the difficulties that dense urban environments present [2]. Additionally, highly detailed maps typically needed for autonomous driving are highly sensitive to environmental changes, and are expensive to obtain in regard to time, money, and resources. Storing these high-fidelity maps on-board the vehicle is unrealistic for all maps in all locations or for environments that change, such as construction areas often found in cities. Data connections could be relied upon to provide the vehicle access to highly detailed maps, however, these connections can be weak, spotty, or non-existent in urban areas. Furthermore, security of autonomous vehicles is also a major concern, as both GPS measurements and maps can be spoofed and/or jammed [3]. Given these challenges, this study investigates alternative architectures and sources of information that may be used for robust navigation of urban roadways.

Currently, most autonomous driving systems use high precision GPS signals to localize within a highly detailed environmental map. Many of the foremost competitors in the autonomous driving industry, such as Google, Uber, and Ford, have entire teams dedicated to obtaining and updating their high-definition (HD) maps, which include everything from lane markings to potholes. These maps are acquired by manually driving vehicles while collecting 360-degree lidar and/or camera data of the environment in which the autonomous vehicle will later drive [4]. This data then undergoes heavy post-processing to form the HD map. Due to the extreme complexity of this task, automakers such as Volkswagen, BMW, and General Motors have relied on third-party services, such as HERE and MobilEye, to provide these highly detailed maps.

To avoid the difficulty of obtaining HD maps in which to localize the vehicle using high precision GPS, research efforts exist in the robotics community to address the problem of navigating without such high-fidelity information sources. On-line *Simultaneous Localization and Mapping* (SLAM) can be applied to autonomous driving to alleviate the need for precise GPS measurements and highly-detailed maps [5], [6]. For example, FAB-MAP, a topological SLAM technique, has been shown to allow for appearance-based navigation [7]. Similarly, SeqSLAM is another SLAM technique that aims to allow for visual navigation despite changing environmental conditions [8]. Although both methods do not rely on GPS measurements or an *a priori* map of the environment, they navigate purely in appearance space and make no attempt to track the vehicle in metric coordinates; in other words, the techniques behave similar to a scene-matching algorithm. Without a way to track the vehicle in metric coordinates, it is impossible to locate the vehicle when it is in between two matched scenes. Therefore, a sufficiently-dense map of images is needed for adequate localization, which causes the algorithms to quickly become computationally expensive and offer sharply diminishing performance as the scale of the environment grows.

Techniques which are less computationally arduous can address the problems of navigating without GPS and a highly detailed map separately. To reduce the vehicle's reliance upon GPS, several methods have been proposed in recent years based on accurate self-localization in mapped environments [9] – [13]. However, these techniques still rely heavily on *a*

*Research supported by NSF GRFP.

All authors are with the Sibley School of Mechanical and Aerospace Engineering, Cornell University, Ithaca, NY 14853, USA {jbc274, mc288}@cornell.edu.

*priori* map information, coming either from lidar or vision data. Similarly, a PosteriorPose algorithm has been shown to keep the navigation solution converged in extended GPS blackouts by augmenting GPS and an inertial navigation system with vision-based measurements of nearby lanes and stop lines referenced to a known map of environmental features [14]. This algorithm retained a converged and accurate position estimate during an 8-minute GPS blackout. To address autonomous navigation without *a priori* map data, GPS-fused SLAM techniques have also been proposed [15], [16]. However, the assumption of consistently receiving these GPS measurement updates is not valid for urban applications, such as in urban canyons like Manhattan and Chicago, and therefore should not be relied upon.

While recent research efforts have been made to face the challenges of driving either in GPS-denied circumstances or without an HD map of the environment, there is considerably less research to simultaneously address both challenges in a computationally-efficient manner. This work aims to explore the extent to which a vehicle can navigate in an urban environment while assuming a varying degree of external information. First, as a worst-case situation, this work studies how far a vehicle can travel with no GPS measurements and no *a priori* map information, other than an initial starting location and measurements from wheel encoders and a compass. Second, this paper explores how far a vehicle can travel with a minimal amount of information, which, for the purposes of this study, comes as a sparse map of landmarks. This results in a navigation solution with much better scalability than many other techniques in the research community. While the focus of this paper is on navigation with limited information, this work could also be used to supplement current navigation systems that use GPS, HD maps, and/or SLAM techniques. This supplemental technique could provide indispensable aid to autonomous vehicles for cases when the navigation system experiences difficulties or its security is threatened; this robustness to infrequent, yet crucial, events is critical for a long-term navigation solution.

## II. SYSTEM ARCHITECTURE

### A. System Overview

The system architecture developed for this study is shown in Fig. 1. This system diagram contains common elements to autonomous driving such as steering and speed controllers, an object tracker, and a path generator. However, the pose estimator and navigation algorithm are updated from their typical form to address the challenges associated with the lack of map information and GPS measurements. The proposed algorithm assumes local sensors allow for the vehicle's real-time control (i.e. staying in a lane). Therefore, precise in-lane localization is not needed for this approach. Rather, high-level localization is provided by the pose estimator, which utilizes only odometry measurements, compass measurements, and sparse map-based measurements, which come from an on-board sparse map of landmarks with corresponding coordinates. This estimator is termed "lightweight" due to the limited amount of sensor measurements it requires. The sparse map-based measurements generated from computer vision methods compare raw camera images to landmark images contained within a sparse map. For the purposes of this study, the map information is assumed to be limited (i.e. no global roadmap, but only a sparse map of images and their corresponding set of coordinates).

The roadway scene includes information such as lane line markings, road signs, traffic lights, and other roadway information that can be extracted from sensor measurements. However due to the focus of this study on navigation and estimation, the roadway scene is assumed to be known. Finally, the roadway scene information, along with the inertial pose estimate, feeds into an intersection navigation algorithm and is used to probabilistically determine the best route to take to reach the goal based on the current limited belief. This high-level navigation scheme is provided by a compass-based navigation control law.

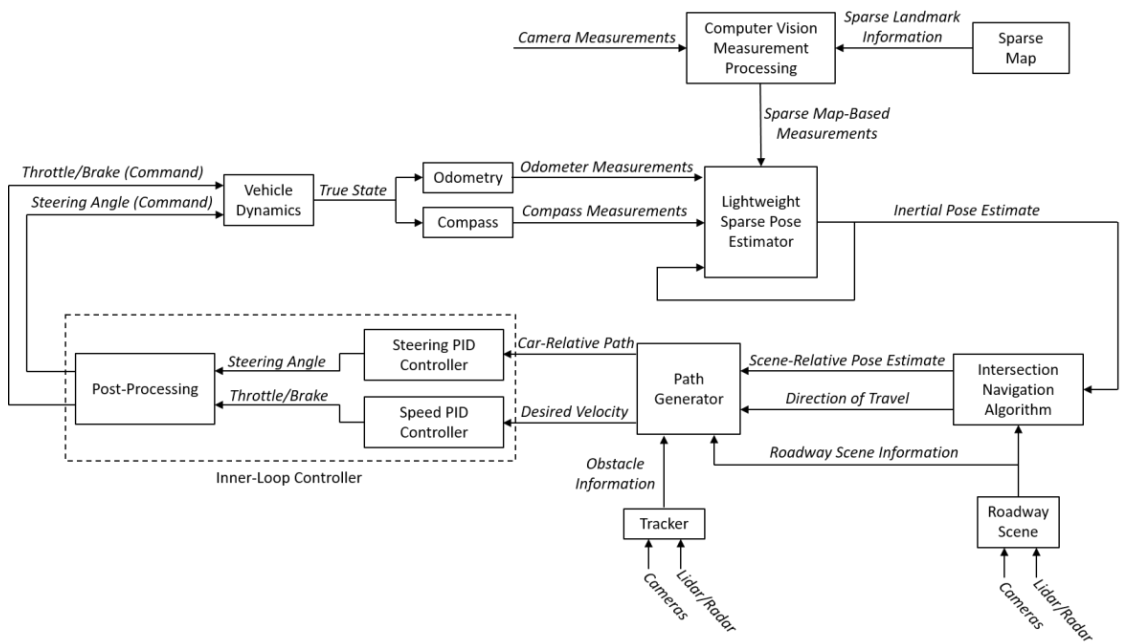

Fig. 1. System architecture for autonomous urban navigation with limited external information.

## B. Lightweight Sparse Pose Estimation

As a first step in this study, a pose estimator is developed assuming no detailed *a priori* map information and no GPS measurements. The pose estimator relies on globally-known start and goal locations of the vehicle, dead reckoning using odometry and compass measurements to estimate the pose of the vehicle within a local frame of reference, and measurement updates to sparse, but known, landmarks. Dead reckoning using odometry measurements alone diverges over time. However, the addition of a compass measurement update, which directly measures the heading of the car, improves the pose estimator accuracy notwithstanding relatively high compass uncertainty. To enable longer driving, the pose estimator is improved using sparse map-based measurement updates. This allows two key questions to be addressed in this work: 1) How far can the vehicle travel during a GPS blackout and with no map information? 2) What level of map landmark sparsity enables the vehicle to successfully navigate a certain distance?

A sparse map of landmarks with corresponding images and coordinates is assumed to be contained within a database on the vehicle. Therefore, as the vehicle travels in the environment and collects camera data, it performs scene detection via computer vision techniques to compare the collected images to the images of landmarks within its database. In this work, ORB feature detection is used to detect and describe features within the local scene of the vehicle and then match to a corresponding urban scene within the database of images on the vehicle reference [17]. When the ORB feature detector obtains a test image that matches an image in the landmark database, the corresponding landmark location is used as a measurement update within the extended Kalman filter (EKF) to refine the pose estimate of the vehicle.

This technique is implemented in simulation as described in section III, and then in experiments as described in section IV. It is noted that the proposed work can make use of any scene detector or computer vision method without loss of generality. However, ORB feature detection was chosen rather than a more sophisticated approach, such as the bag-of-words model used in FAB-MAP, due to its ease of implementation and the fact that it does not need to be extensively trained and tuned. Furthermore, by tracking the vehicle's position in metric coordinates in between landmark detections via dead reckoning, a mask can be generated to disregard all landmarks beyond the 2-sigma uncertainty ellipse around the vehicle. This caused the ORB landmark detector to be very computationally efficient – as it did not need to compare the current test image to all the images in the database, but only to the database images closest to the vehicle – and it also improved performance, as it essentially eliminated all false positives for the experiments described in section IV.

## C. Intersection Navigation Algorithm

Navigating to a desired location in an urban environment without the availability of GPS or detailed map information is a difficult challenge. Certain assumptions must be made for this problem to become feasible. First, the coordinates of the start and end points are assumed to be known within a defined uncertainty. Next, the vehicle is assumed to be equipped with a suite of sensors that allows it to detect a variety of common roadway objects such as road signs, traffic lights, cyclists, pedestrians, other vehicles, and lane markings. The ability to detect at this level is important because it allows the vehicle to avoid obstacles, stay on the road by following lane lines, and detect when it is approaching an intersection; however, these measurements are not used for pose estimation. Given the readily available lidar units, radar units, cameras, and computer vision technology on the market, this is a valid assumption to make. For instance, MobilEye's vision-based advanced driver assistance system can detect lane lines, other vehicles, pedestrians, and various traffic signs [18]. Many autonomous vehicles have this detection capability. However, since this study aims to address novel pose estimation and navigation techniques, these detection capabilities were not deemed necessary to implement for this research.

Assuming such a suite of sensors is available, the low-level navigation problem becomes feasible, and a higher-level navigation problem can be formulated. Navigation to a desired location from a given starting point follows two basic principles in this framework. First, as the vehicle approaches an intersection and is faced with a decision of its direction of travel (e.g. continue straight, turn left, turn right, etc.), the vehicle minimizes the difference in angle between the heading of the vehicle and the direction to the goal after the proposed intersection decision. Second, the vehicle retains a list of intersections that it has already visited and the corresponding decision made at each intersection; note that this list of visited intersections is based on the vehicle's pose at the time an intersection is detected, which is subject to uncertainty and will become less reliable as the pose uncertainty increases. As the vehicle approaches a previously-visited intersection, the navigation algorithm applies a penalty to the prior intersection decision, thus, making it less probable to make the same decision as before. The second principle is similar to Tabu search [19], as it relaxes the basic rule of the algorithm and discourages the search to return to previously-visited solutions to avoid getting stuck in suboptimal regions. The decision rule is general to any type of intersection (e.g. different number of roads at different angles), although this study focuses on gridded roadways due to the intended application to urban environments.

These two principles for the navigation algorithm are summarized in the following optimization problem:

$$\theta^* = \underset{i \in N}{\operatorname{argmin}}[|\theta_i^+ - \varphi_i| + \gamma_i], \qquad (1)$$

where $\theta^*$ is the intersection decision, or direction of travel after the intersection, $i$ represents the index of turning options at an intersection (e.g. continue straight, turn left, turn right, etc.), and $N$ is the total number of turning options at a given intersection. The direction of travel of the vehicle after the intersection is denoted by $\theta_i^+$. The direction from the vehicle to the goal after the intersection is represented by $\varphi_i$. The penalty of making the same decision at the same intersection as before is given by $\gamma_i$, which increases incrementally each time the same decision is made at the same intersection.

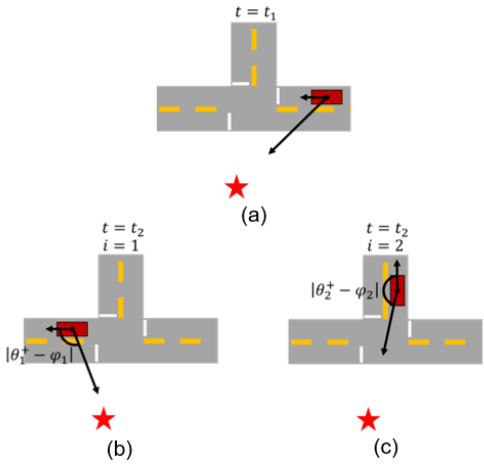

Fig. 2. Illustration of an autonomous vehicle navigating through a 3-way intersection. The goal is denoted by the red star. (a) The vehicle approaching the intersection. (b) The vehicle after continuing straight through the intersection (decision $i = 1$). (c) The vehicle after turning right through the intersection (decision $i = 2$). The navigation algorithm would choose to continue straight through the intersection in (b), assuming no previous visits to this intersection.

Fig. 2 shows an example of an intersection decision to better understand the navigation algorithm in (1). In this example, the vehicle navigates through a 3-way intersection. The navigation algorithm assumes that the vehicle can detect the type of intersection that it is approaching (e.g. 3-way intersection) based on the measurements of its on-board sensors. With this information, the algorithm computes the angle between the vehicle's heading and the direction to the goal for each possible route. After including any penalties for the decisions made at prior visits, the route associated with the minimum value is selected. In this example, assuming no penalties, the algorithm would direct the vehicle to go straight through the intersection.

## III. SIMULATION RESULTS

A simulator was developed to test the feasibility of the proposed pose estimation and navigation techniques. The simulator models the vehicle's dynamics using a four-state bicycle model [20] and simulates its motion through a randomly-generated city road network. A predictive state model of the vehicle is given as

$$x_{k+1} = f(x_k, a_k, \varphi_k, \delta t) \quad (2)$$

$$\begin{bmatrix} x_{k+1} \\ y_{k+1} \\ \theta_{k+1} \\ v_{k+1} \end{bmatrix} = \begin{bmatrix} x_k + \Delta x \cos(\theta_k) - \Delta y \sin(\theta_k) \\ y_k + \Delta x \sin(\theta_k) + \Delta y \cos(\theta_k) \\ \theta_k + \Delta \theta \\ v_k + a_k \delta t \end{bmatrix}. \quad (3)$$

In the equations above, $x_k$ is the vehicle state at time $k$, which consists of four states: the position of the center of the vehicle's rear axle ($x_k$ and $y_k$), the heading of the vehicle ($\theta_k$), and the speed of the vehicle ($v_k$). The control inputs to the vehicle model are acceleration ($a_k$) and steering angle ($\varphi_k$). The prediction time step is given as $\delta t$. In addition, the $\Delta$ terms are defined as

$$\Delta\theta = \frac{d_k}{\rho_k}, \quad \Delta x = \rho_k \sin(\Delta\theta), \quad \Delta y = \rho_k (1 - \cos(\Delta\theta)),$$

where

$$\rho_k = \frac{l}{\varphi_k}, \quad d_k = \frac{1}{2} a_k \delta t^2 + v_k \delta t.$$

The parameter $l$ is the length between the front and rear axles, $\rho_k$ represents the radius of curvature for the vehicle, and $d_k$ represents the distance the vehicle travels over $\delta t$.

During the vehicle's travel, simulated odometry, compass, and vision measurements are used to update the vehicle pose estimate. Steering and speed proportional-integral-derivative (PID) controllers are used to allow the vehicle to stay on a specified path chosen by the navigation algorithm. A randomly gridded map was generated for each test in this Monte Carlo study. These maps ranged in size from 1 sq. km to 100 sq. km, and the block size ranged from 50 meters to 300 meters. Dead end and one-way roads were randomly scattered into the map. Random start and end locations were chosen. Finally, for the landmark detection study, landmarks were scattered into the map at random locations according to the given landmark density.

### A. Monte Carlo Range Tests without Map Information

Monte Carlo studies were conducted in simulation to determine the maximum distance an autonomous vehicle can travel without receiving external pose measurements (i.e. no GPS or landmark measurements) before it becomes lost. Without the use of a map, local sensors are assumed to enable the vehicle to stay on the road. The limiting factor in reaching the goal is the navigation algorithm's ability to differentiate between each intersection. At a sufficiently high position uncertainty level, it becomes ambiguous which intersection the vehicle is approaching and the high-level navigation breaks down. The vehicle is assumed to be lost when the major axis of the 2-sigma uncertainty ellipse of the vehicle's x-y location grows to exceed 100 meters, which is approximately the average size of a Manhattan city block [21]. A 2-sigma uncertainty ellipse, rather than a 1-sigma uncertainty ellipse, is used to provide a high confidence region for the location of the vehicle. While the uncertainty is below this threshold, the vehicle is considered sufficiently localized for the high-level navigation algorithm to function.

This study assumes a conservative odometer measurement uncertainty common in cars and ABS braking systems. This uncertainty considers wheel slip and rotary encoder discretization errors, which have a maximum error of ±1/2 of the angular rotation between two successive bits [22]. For compass measurements, interference from the large amount of electrical hardware found on an autonomous vehicle can result in high uncertainty. Therefore, three values for the 2-sigma compass uncertainty are studied here: ±10°, ±20°, and ±30°. For comparison, an additional set of simulations were performed in which no compass measurements were used. A total of 4,000 simulations were conducted, with each simulation using a new random map, and new random start and end points. Fig. 3 plots the 2-sigma ellipse major axis as a function of distance traveled for each trial in all three compass uncertainty cases, as well as the case with no compass measurements.

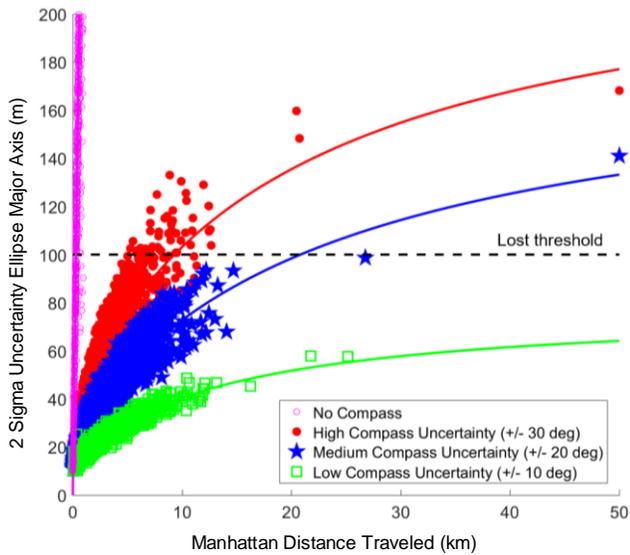

Fig. 3. Monte Carlo results to determine the distance an autonomous vehicle can travel without external pose measurement updates before getting lost, which is defined by the threshold indicated by the horizontal dashed line; note that this figure shows the Manhattan distance traveled by the car.

The results for each set of tests follow a predictable trend: the major axis of the 2-sigma uncertainty ellipse increases quickly when the ellipse is small and then grows more gradually as the area of the ellipse becomes large. For a 2-sigma compass uncertainty of ±10°, a distance at which the vehicle becomes lost was not found. For a 2-sigma compass uncertainty of ±20°, the vehicle could travel 20.8 km on average before it became lost. For a 2-sigma compass uncertainty of ±30°, the vehicle could travel 9.3 km on average before it became lost. Finally, for the case with no compass measurements, the vehicle could only travel 300 meters on average before it became lost. In this case, the 2-sigma uncertainty grew very rapidly since only wheel encoders were being used to estimate the position of the car. Therefore, as the vehicle began to drive, the initial uncertainty in the heading of the vehicle quickly resulted in a large amount of lateral uncertainty in the position of the vehicle.

The main source of variation in this study relates to the number of turns that the vehicle made during its travel. It is expected that the final uncertainty ellipse major axis is larger for a vehicle driving along a straight road compared to a vehicle driving the same distance while taking many turns. This is seen in Fig. 3, as most of the data points above the fitted curve resulted from tests in which the vehicle took few turns, while the majority of the data points below the fitted curve resulted from tests in which the vehicle took many turns. This is because the ellipse grows predominantly along only one axis in the case of driving straight and along both axes in the case involving many turns, and therefore the major axis of the 2-sigma uncertainty ellipse grows at a faster rate for straight driving.

### B. Monte Carlo Landmark Detection Study

A Monte Carlo study was performed to determine the effect that map-based measurements from sparse landmarks have on the pose estimation problem and subsequent navigation. The parameters of this study were the same as the prior Monte Carlo range tests, except this study utilizes a sparse map of landmarks with corresponding coordinates known *a priori* to allow the vehicle to perform pose updates as it drives. A camera takes measurements of a scene, and attempts to correlate a detection with a sparse map of locations. If a positive detection is made, a map-based pose measurement update is performed using an uncertain location.

Furthermore, this study explores three variants of the heading-based navigation function in (1). The first method, termed the straight to goal method, attempts to drive straight to the goal and does not actively seek out landmarks (i.e. fortuitous landmark measurement updates). The second method, called the landmark to landmark method, seeks out the closest landmark to the vehicle that also moves the vehicle closer to the goal. The third approach is the hybrid method, where the vehicle attempts to drive straight to the goal until its pose uncertainty exceeds a specified threshold, at which point the vehicle then seeks out landmarks to improve its pose estimate. For this study, the specified threshold is 50 meters for the major axis of the 2-sigma position uncertainty ellipse.

In addition to the navigation method, the effects of landmark density and landmark detection rate are also studied. The simulated landmark detection rate is implemented by disregarding a specified percentage of the landmark detections. A constant 2-sigma compass uncertainty of ±30° is assumed, which is the most conservative value from the Monte Carlo range study with no map information. With 3 different navigation methods, 5 different landmark densities, and 5 different landmark detection rates being considered, a total of 75 different combinations of test conditions are studied. For each combination, 700 simulation trials were conducted to find the success rate as a function of Euclidean distance from starting point to goal. Success rate is defined as the rate at which the vehicle reaches the goal without its 2-sigma uncertainty ellipse major axis exceeding the lost threshold of 100 meters.

Fig. 4 summarizes the performance results from more than 50,000 simulations in this Monte Carlo study. Note that Fig. 4 shows range as Euclidean distance, where Fig. 3 shows Manhattan distance traveled.

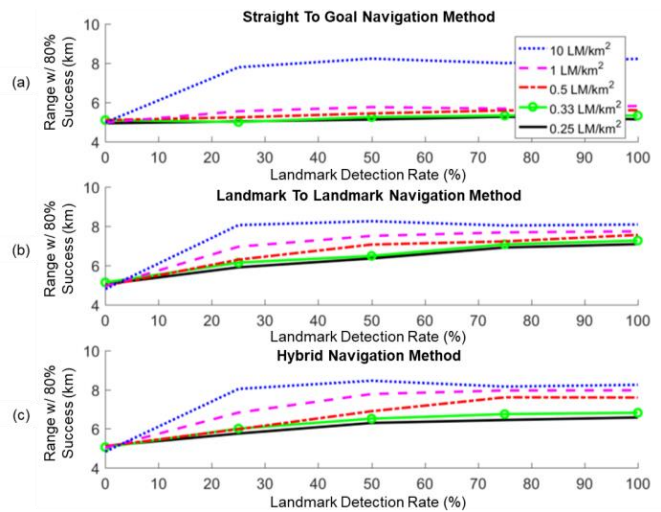

Fig. 4. Monte Carlo landmark detection study results for (a) straight to goal navigation, (b) landmark to landmark navigation, and (c) hybrid navigation; note that this figure shows the range as Euclidean distance traveled by the car, rather than Manhattan distance.

Due to the different navigation methods resulting in different routes to the goal, the Euclidean distance from start point to goal, instead of the total Manhattan distance traveled, is used in Fig. 4 for a fair comparison between navigation methods. The performance defined for this study is an 80 percent success rate of reaching the goal; this success rate is chosen for easy comparison with the experimental results shown in section IV. For each test condition in this study, the range of the vehicle at a specified success rate gives a good indication of the overall success of the test condition. It is expected that the range and overall success of the navigation method increase as the landmark density and landmark detection rate increase. In general, this trend is seen in the results. For dense landmark maps, a clear upward trend can be seen in the data, with the trend becoming subtler as the density becomes sparser. The trend is also subtler for the straight to goal navigation method, since the vehicle is not actively seeking out landmarks and therefore receives far fewer landmark measurement updates. Overall, Fig. 4 allows for the expected range with a given success rate to be obtained for all combinations of navigation methods, landmark densities, and landmark detection rates.

Fig. 4 also gives insight to the effectiveness of each navigation method. The straight to goal navigation method, where the vehicle ignores the landmarks and attempts to drive straight to the goal, performs noticeably worse compared to the other two methods, which perform similarly in terms of robustness (i.e. the reliability of the vehicle to reach the goal for given test conditions is similar). Excluding the 10 landmarks per sq. km case (in which many landmarks are reached regardless of the navigation method), the range of the straight to goal navigation method for all other test parameters is typically 1 to 2 km less than the corresponding range for the other two navigation methods. However, the navigation methods that actively seek out landmarks (the landmark to landmark and hybrid navigation methods) increase the average distance traveled for the vehicle. The landmark to landmark navigation method causes the vehicle to travel 31 percent further on average compared to the straight to goal navigation method. Similarly, the hybrid navigation method causes the vehicle to travel 15 percent further on average compared to the straight to goal navigation method. Therefore, in general, there is a tradeoff between distance traveled and robustness for the navigation methods that sought out landmarks to update the vehicle's pose estimate.

## IV. EXPERIMENTAL RESULTS

To verify the simulation results and understand the maturity of the theory, the proposed estimation and navigation techniques were implemented on a 2007 Chevrolet Tahoe and tested in downtown Ithaca, NY. A picture of the test vehicle is shown in Fig. 7. Due to the current traffic laws in the state of New York, the pose estimation and navigation techniques were implemented and used to guide a human driving the car, directing the driver where to go. In other words, the driver was merely used to act as the inner-loop controller and keep the vehicle on the road. A video of the results accompanies this paper.

The odometry measurements for these field tests were obtained from the vehicle's wheel encoder measurements on the CAN bus. A low-cost compass was installed in the vehicle to receive heading measurements. Due to the large amount of electromagnetic interference in the car, the compass' 2-sigma uncertainty was empirically determined to be approximately ±25°. However, this uncertainty is dependent on the location of the vehicle and can improve or worsen based on its location within the city. Finally, a Point Grey Ladybug3 360-degree camera was used to capture images for landmark detection.

### A. Range Tests without Map Information

The first set of experiments aimed to verify the results from the Monte Carlo range study with no map information in Fig. 3. For these experiments, the vehicle was driven for a specified amount of time and the final position uncertainty from the pose estimator was recorded. Fig. 5 shows the results from 33 range test experiments with no map information overlaid with the Monte Carlo simulation results.

Results show that the vehicle can travel nearly 10 km with no map information before becoming lost. Given a 2-sigma compass uncertainty of approximately ±25°, the experimental results are expected to fall between the high compass uncertainty and medium compass uncertainty data points. As shown in Fig. 5, this is typically the case, reinforcing the simulation results. However, as the distance traveled increases, the experimental data align more closely with the high compass uncertainty data points. This is likely due to locations where the compass experienced very high magnetic interference, which was seen during calibration tests. As the distance traveled by the vehicle increases, the more likely it is to drive through one of the regions with high magnetic interference, leading to a higher average compass uncertainty. In the supplemental video, the vehicle can be seen to travel through one of these high interference regions at 2:16.

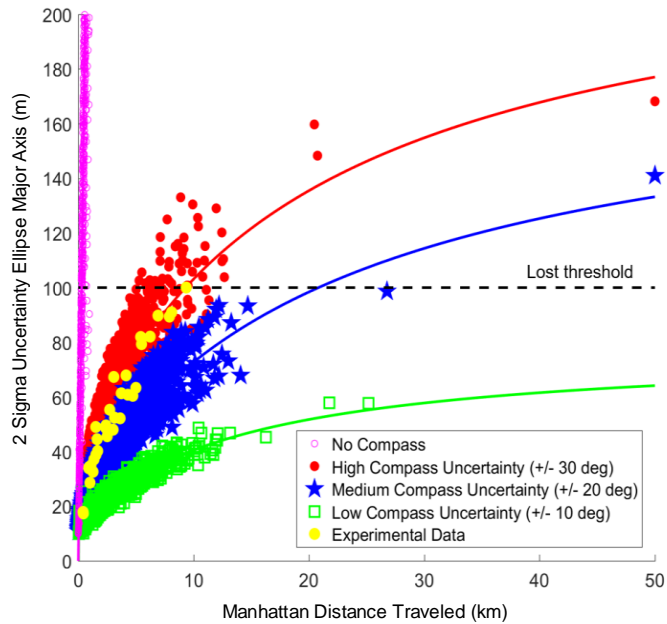

Fig. 5. Experimental results shown overlaid with simulated Monte Carlo results to verify the Manhattan distance an autonomous vehicle can travel without external pose measurement updates before getting lost. Thirty-three field tests were performed for this study.

## B. Landmark Detection Tests

To verify the Monte Carlo landmark detection results, an ORB landmark detector was developed and implemented on the vehicle to augment the existing pose estimator. This landmark detector had an average detection rate of approximately 60 percent. Furthermore, approximately 100 images from Ithaca's downtown intersections were obtained to populate the vehicle's landmark database; however, only a few of these images were used during the experiments due to the specific landmark density that was chosen to be tested. Finally, the landmark to landmark navigation method was used in these tests due to its high robustness.

For each test, the vehicle's navigation algorithm guided it to a randomly chosen goal location while relying on the sparse pose estimator for localization. Once the goal was reached, a new random goal was spawned and the vehicle then proceeded to drive to it. This process was repeated until the vehicle became lost and the navigation algorithm broke down or the experiment exceeded 75 minutes. For each goal generated, at most one landmark would be randomly placed within the map. This landmark would then be cleared when the vehicle reached its corresponding goal. Therefore, no more than one goal and one landmark were on the map at a time. This approach of traveling to many subsequent goals, as opposed to one goal, needed to be taken due to the small size of Ithaca's downtown (approximately 1 sq. km). Simulations show that the method of subsequently traveling to many short-range goals produce analogous results compared to traveling to one long-range goal. The supplemental video shows how these experiments were performed.

Ten experiments were conducted with an average landmark density of 0.55 landmarks per sq. km and an average landmark detection rate of 60 percent; the number of tests was chosen based on the time required to run each trial (approximately 1 hour). In 8 of these 10 tests, the vehicle successfully reached a final goal with a Euclidean distance of at least 6.9 km from the starting point. This result is plotted in Fig. 6 along with the Monte Carlo simulation results.

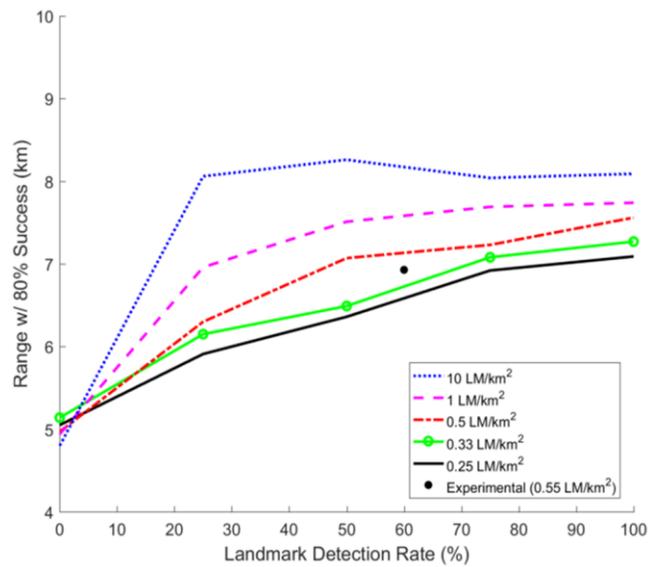

Fig. 6. Experimental result shown with simulated Monte Carlo landmark detection study results. Ten field tests were conducted using the landmark to landmark navigation method to verify the simulation results. Note that this figure shows the range as Euclidean distance traveled by the car, rather than Manhattan distance.

When incorporating a sparse landmark-based map, the vehicle can reliably travel to a much further goal, as compared to the tests without map information. In many of these experiments, the vehicle could drive a Manhattan distance of 20 km (more than twice the distance traveled in the tests with no map information). The experiment showed that the vehicle could drive nearly 10 km between landmark measurement updates without getting lost. Therefore, the experiments demonstrated that, if the landmark density and detection rate were high enough to guarantee a landmark measurement update at least every 10 km traveled by the vehicle, then the vehicle could travel indefinitely without getting lost. This was seen in 3 tests, as the vehicle received numerous landmark measurement updates during its travel; however, these tests were eventually cut short after 75 minutes of driving.

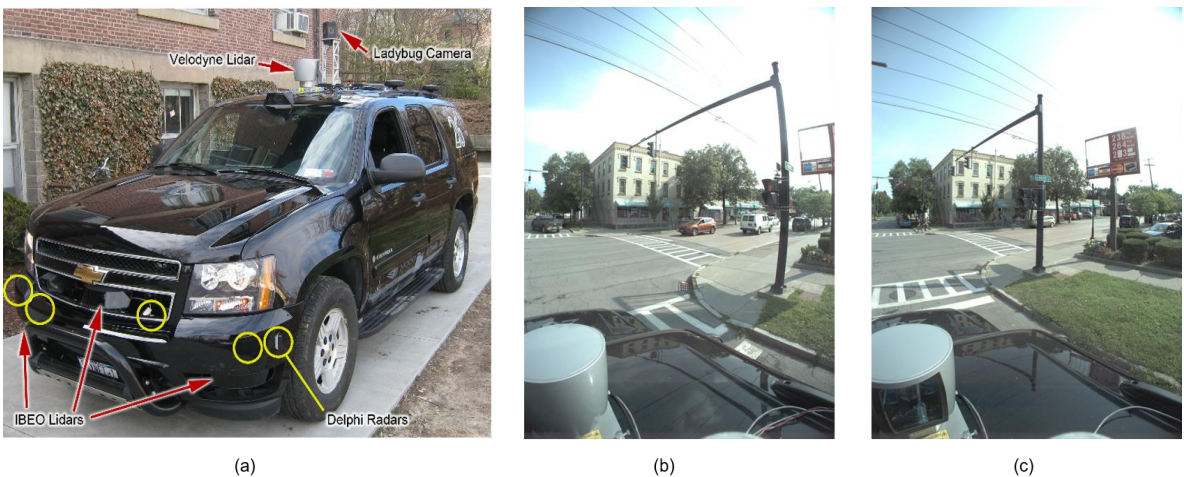

Fig. 7. (a) Test vehicle with notable sensors indicated; note only the Ladybug camera was used in this study, the lidar and radar units were not used. (b) Test image of a landmark from the Ladybug camera during the experiments. (c) Database image of the landmark corresponding to the image in (b). These images produced a match and resulted in a landmark measurement update despite being taken on different days under different conditions. Note the resolution of the original images is 1232 x 1616

While the average experimental results match the simulation results relatively well, there is some variation. A total of 10 trials is small compared to the number of simulation trials. It is expected that the experimental result would more closely reflect the simulated results with additional tests. Given the large amount of time needed to perform each test, 10 experiments were deemed sufficient to demonstrate the capabilities of the proposed techniques. Additionally, the landmark detection rate varied from test to test. This high uncertainty in the detection rate could also explain why the experimental result is lower than expected in Fig. 6. The detection rate was dependent on many environmental conditions including the prominence of the features at each intersection, the weather, and the traffic. For instance, the vehicle would be less likely to receive a landmark measurement update if the landmark did not have interesting features (e.g. an open field or empty parking lot), or if the weather or traffic was very different from the database images. An example of traffic leading to a missed landmark measurement update can be seen in the supplemental video at 1:30, where a truck parked too close to an intersection caused the test vehicle to move into the opposite lane. The resulting image taken at that landmark was from an angle in the opposite lane, resulting in a missed detection. Fig. 7 shows the test vehicle, as well as a database image and test image of a landmark taken during testing.

## V. Conclusions

A novel system architecture is presented to address the problem of autonomous driving within an urban environment when reliable GPS measurements and map information is limited. This paper proposes a pose estimation method that utilizes odometry, compass, and sparse map-based measurements to estimate the pose of the vehicle as it autonomously navigates the roadways with limited map information and GPS measurements. This study also uses a simulator to study key parameters of the navigation and pose estimation algorithms within the proposed system architecture. Monte Carlo studies using this simulator provide evidence to resolve key issues concerning navigating without GPS or detailed map information. These studies show the distance a vehicle can travel with no GPS or map information, as well as the relationship between the range of the vehicle, navigation method, landmark density, and landmark detection rate. Experimental results verify the simulation results within a small amount of deviation, as they produce a minimum range of 6.9 km for the given success rate, navigation method, landmark detection rate, and landmark density.

## Acknowledgment

This work was supported by the NSF GRFP grant DGE-1144153.